\documentclass[runningheads]{llncs}
\usepackage[T1]{fontenc}
\usepackage{graphicx}
\usepackage[disable]{todonotes}
\usepackage{multirow}
\usepackage{amsfonts}

\begin{document}
\title{Calibrate to Interpret}
%\author{Anonymous Authors}
\author{Gregory Scafarto\inst{1}\orcidID{0000-0002-6544-2358} \and
 Nicolas Posocco\inst{1}\orcidID{0000-0002-1795-6039} \and
 Antoine Bonnefoy \inst{1}\orcidID{0000-0003-1720-7326}}

\authorrunning{G. Scafarto et al.} % G. Scafarto et al.
\institute{ EURA NOVA, Marseille, France \and
\email{firstname.lastname@euranova.eu}
}
\maketitle              % typeset the header of the contribution
\begin{abstract}
  Trustworthy machine learning is driving a large number of ML community works in order to improve ML acceptance and adoption. 
  The main aspect of trustworthy machine learning are the followings: fairness, uncertainty, robustness, explainability and formal guaranties.
  Each of these individual domains gains the ML community interest, visible by the number of related publications. However few works tackle the interconnection between these fields.
  In this paper we show a first link between uncertainty and explainability, by studying the relation between calibration and interpretation.
  As the calibration of a given model changes the way it scores samples, and interpretation approaches often rely on these scores, it seems safe to assume that the confidence-calibration of a model interacts with our ability to interpret such model. In this paper, we show, in the context of networks trained on image classification tasks, to what extent interpretations are sensitive to confidence-calibration.
  It leads us to suggest a simple practice to improve the interpretation outcomes : \emph{Calibrate to Interpret}. 

  %Within the past few years, many research works have tried to increase the adoption of machine learning models. Both the knowledge of uncertainty in predictions and the ability to explain such predictions can be used to that end, since they provide insights which strengthen users' trust. As the calibration of a given model changes the way it scores samples, and interpretability approaches often rely on these scores, it seems safe to assume that the confidence-calibration of a model interacts with our ability to interpret such model. In this paper, we show, in the context of networks trained on image classification tasks, to what extent interpretations are sensitive to confidence-calibration, and investigate the reasons why.
\end{abstract}
\section{Introduction}
Despite being state of the art on many tasks, deep neural networks (DNNs) are still considered as black boxes which is problematic in contexts where decision making is critical.
In order to ensure that a model can be safely used, one needs to have access to the uncertainty over its predictions, and to understand what drives those predictions. 
Both aspects, namely uncertainty and explainability, are frequently tackled via the use of post-hoc calibration and local interpretation respectively.
The current work studies the interaction between these two central aspects of trustworthy machine learning.
% Being able to understand the behaviour of a model can also be valuable in many cases such as debugging their behavior, certifying that they have some desirable generalization properties, and that their predictions are fair \cite{Mythos_Lipton}.

\begin{figure}[t]
    \includegraphics[width=0.94\textwidth]{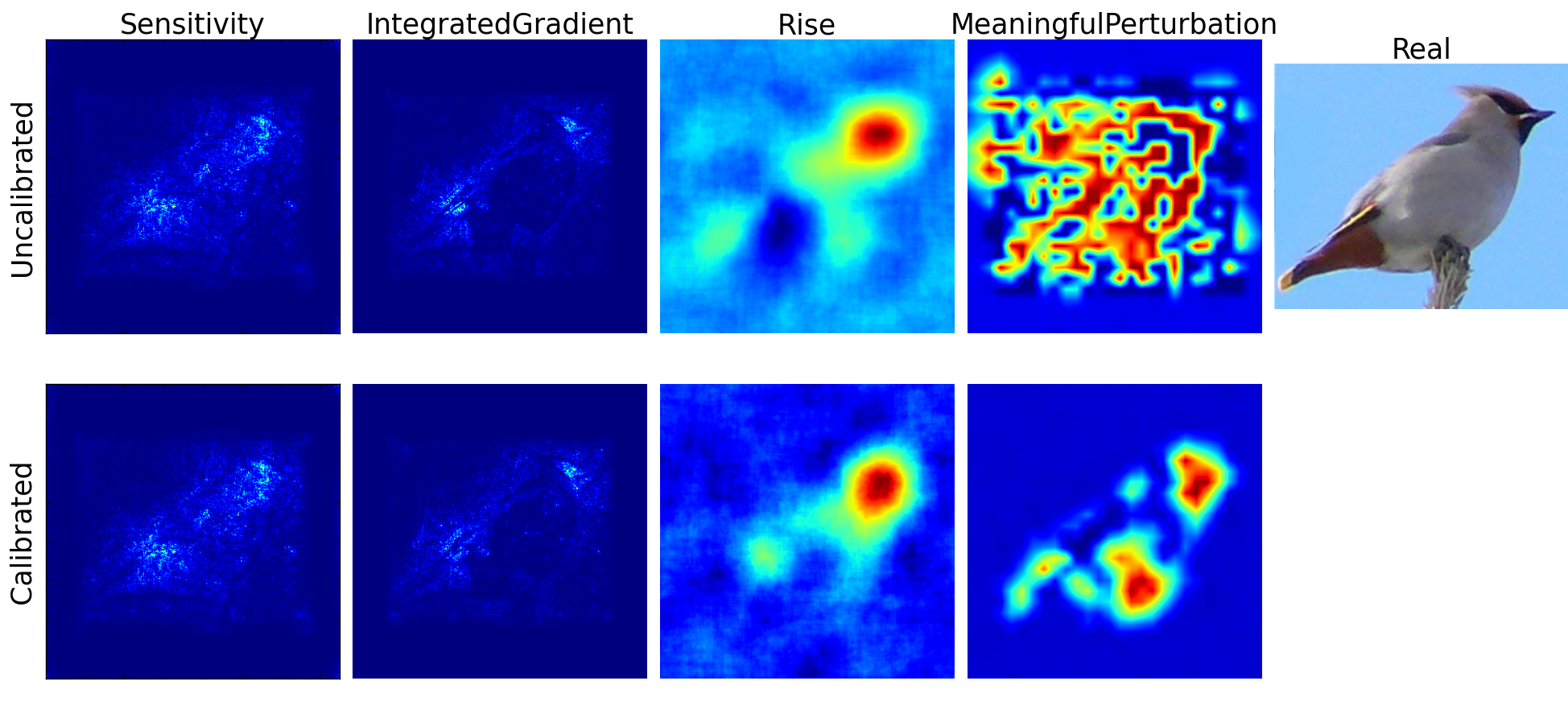}
    \centering
    \caption{Visual comparison of saliency changes due to calibration}
    \label{fig:saliency-example}
\end{figure}

There are numerous ways to interpret a model's behaviour, depending on what one has access to (internal model structure, training phase, ...).
This work focuses on methods which interpret the decisions of already trained image classifiers, considering model-aware as well as model-agnostic local interpretation methods, following the definitions given in \cite{ribeiro2016why}, the former have access to the internal structure of the model while the latter only relies on the predictions scores. 
These methods provide a saliency map highlighting important pixels for each prediction.
%They are the most used explanation approaches to provide information about Deep Neural networks' decision, and integrate seamlessly with the exploitation of trained models.  
% from white-box approaches which use the understandable structure of a simple model to black-box approaches, which have access to the internals of the model -model-aware (ref)- or not -model-agnostic (ref)- . Some methods even require training similar models to build interpretations, like (ref). Interpretations can be global -high-level understanding of the dynamics of the model- or local -understanding of each prediction- . 
Regarding predictions uncertainty, post-hoc calibration methods adapt models so that their scores are consistent with conditional probabilities related to the predicted class \cite{kull2019dircal}. It is a convenient way to tackle the overconfidence of modern neural networks~\cite{pmlr-v70-guo17a,Fooled_Nguyen}. 

This focus was made to fit the flexible context in which calibration and interpretation are handled after training the model. Since both post-hoc calibration and local interpretation are linked to the scores associated with the predicted class, 
we address their interdependence via the following questions:
Does calibration impact the saliency maps obtained as interpretations?
Do modifications, if any, improve the faithfulness of interpretation methods?
Are saliency maps with calibrated models more human-friendly?

% We highlight a relationship between calibration and interpretability and show that calibration can be used to improve the produced interpretations

Our empirical evaluations highlights that there is indeed a positive interaction between the calibration of a model and its interpretability by enabling some widely used interpretation methods to work more efficiently on it. 
This impact is visible in terms of faithfulness, visual-coherence of the saliency maps, and their robustness most notably for model-agnostic approaches like Meaningful Perturbation~\cite{8237633}. 
%This impact is particularly strong for model-agnostic approaches. 
Examples of saliency maps produced by various methods before and after the calibrations are presented in Figure~\ref{fig:saliency-example}.

After positioning our work in Section~\ref{sec:rel-works}, we introduce the problems of calibration and local interpretation, as well as the methods used in the literature to tackle them in Section~\ref{sec:prob-stat}. Section~\ref{sec:empirical-study} then describes the conducted experiments and reports their outcomes. Finally results are discussed and we mention some future works before concluding in Section~\ref{sec:concl}\footnote{The code is available at : https://github.com/euranova/calibrate\_to\_interpret}.

\section{Related Works}
\label{sec:rel-works}

% \todo{Following preceding works which linked robustness and calibration \cite{Robtuness_Interp_Slavin,tsipras2018robustness,robutness}, data augmentation and interpretation \cite{kim2020puzzle} or calibration and fairness \cite{pleiss2017fairness}, we study the interaction between two aspects of trustworthy machine learning: uncertainty and explanation via the empirical analysis of the interaction between post-hoc calibration and local interpretation.}

Trustworthy machine learning has recently gained interest, with the development of modern uncertainty quantification, explainability, robustness, fairness and formal guaranties.
%Among these aspects there is uncertainty that aim at characterising the epistemic or aleatoric uncertainty; and robustness to small noise or adversarial samples; fairness to equally treat sensitive features regardless of the biases contained in the training data; explainability and interpretability to extract knowledge from the model and its prediction to make the model decision process more accessible to humans.  
However few publications have investigated links between these aspects.
Among these we can mention attempts to link: robustness and calibration \cite{Robtuness_Interp_Slavin,robutness,tsipras2018robustness} showing that models which are robust to adversarial attacks are more interpretable; data augmentation, calibration and interpretation showing that MixUp data augmentation greatly impacts the calibration of learnt models \cite{thulasidasan2019mixup} and existing saliency methods being used to improve the MixUp procedure itself \cite{kim2020puzzle}; or calibration and fairness \cite{pleiss2017fairness}, showing the incompatibility between most of fairness definitions and calibration.
Following these works, we study the interaction between two aspects of trustworthy machine learning: uncertainty and explainability via the empirical analysis of the interaction between post-hoc calibration and local interpretation.

\section{Problem Statement and other Related Works}
\label{sec:prob-stat}
Let $x \! \in \! \mathbb{R}^ {H \times W \times C}$ be an input image and $x_{i}$ be its features. A model $F$ is trained to classify sample images among $C$ classes. 
It maps each input $x$ to a logit vector $L$ which is then converted, generally through a softmax function, into an output vector $F(x) \in [0,1]^{C}$ so that $\sum_{c=1}^{C} F(x)_{c}=1$. 
The decision associated with such prediction is $ y = \arg\!\max_{c}{F(x)_{c}}$.  %\!\in\!\left\{x_{0} \ldots x_{H*W*C}\right\}

\subsection{Calibration}

\begin{figure*}
    \centering
    {\includegraphics[width=0.32\textwidth]{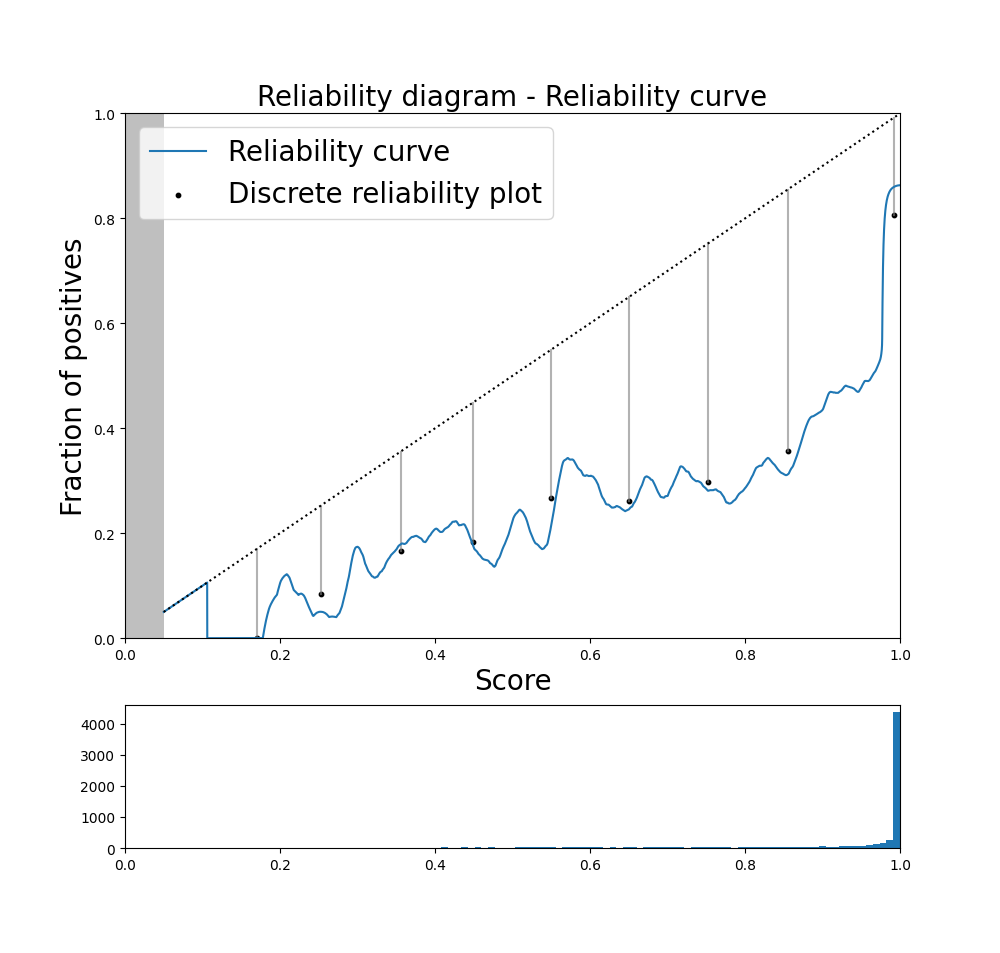}} 
    {\includegraphics[width=0.32\textwidth]{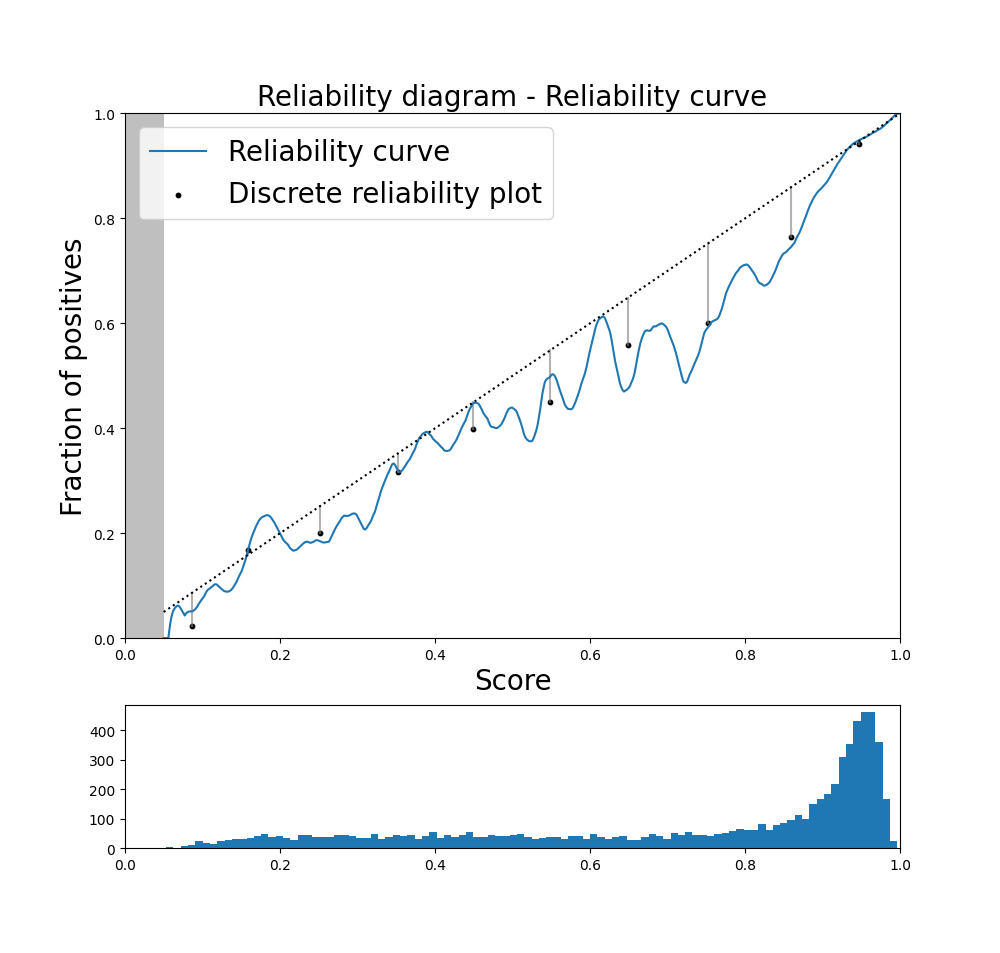}} 
    {\includegraphics[width=0.32\textwidth]{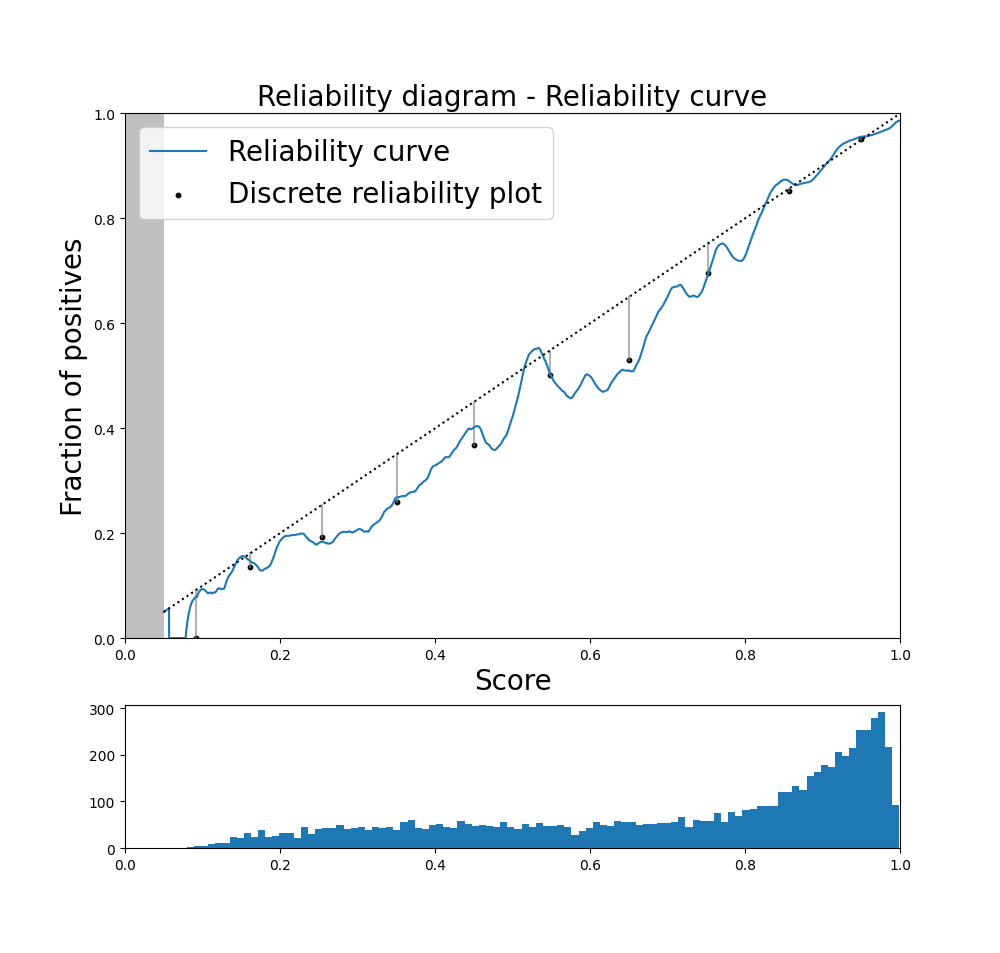}}
    \caption{Confidence reliability plots and curves for VGG16 trained on CIFAR100, when uncalibrated (left, $ECE_{conf}~=~0.2$), calibrated using Temperature Scaling (center, $ECE_{conf} = 0.04$) and calibrated with Dirichlet calibration (right, $ECE_{conf} = 0.05$)}
    \label{calib}
\end{figure*}

Many critical applications motivate the evaluation of confidence-calibration, which indicates how confident a model is in the class it predicts. It is well-known that modern neural networks tend to be miscalibrated, generally overconfident on their predictions \cite{pmlr-v70-guo17a,Fooled_Nguyen}. 
Thus some works have been done both on the evaluation of calibration \cite{Pakdaman_Naeini_Cooper_Hauskrecht_2015} and the improvement of such calibration for a given classifier, would it be during training \cite{pmlr-v80-kumar18a} or as a post-processing step \cite{Beta_Kull2017,kull2019dircal,Platt99probabilisticoutputs}.

As local interpretation allows to interpret each specific predictions, we focused our work on its interaction with confidence-calibration, i.e. the calibration of scores for the predicted class $argmax(F(x))$.

A model is said to be well confidence-calibrated if:
\begin{equation}
\forall s \in [0,1],
P\left(y=z \mid F(x)_{z}=s\right)=s
\end{equation}
with $z = argmax(F(x))$

This notion can be quantified with estimates of the Expected  Calibration  Error (ECE), defined in the confidence-calibration setting as :
\begin{equation}
ECE^{conf}= ECE^{conf}= \mathbb{E}_{max(F(X))} \left [ \left| \mathbb{P}(y=z \mid F(x)_{z}=s)-s \right | \right ]
\end{equation}

%As we are interested in assessing how calibration interacts with local interpretation, which focuses on interpreting each specific predictions, we focus our work on confidence-calibration, i.e. the calibration of scores for the predicted class $argmax(F(x))$. 
In order to calibrate our models, we focus on post-hoc calibration, which can be applied to already trained models and relies exclusively on the scores given by the model. These very nice properties make these approaches plug-and-play, which justifies their popularity. We use in this paper the following techniques, which impact is illustrated in Figure~\ref{calib}.

\paragraph{Temperature scaling}
If $L$ is the logits vector output of the classifier before softmax activation $\sigma$, calibrated scores are given by $F(x)_{c}^{cal} = \sigma(L T^{-1})_c$ \cite{Platt99probabilisticoutputs}.
% \begin{equation*}
%\frac{e^{\frac{L_c}{T}}}{\sum_{j=1}^{C} e^{\frac{L_j}{T}}}%
% \end{equation*}
The temperature scaler $T$ is fit to maximize the likelihood on a holdout set.
Its role is to smooth or sharpen predicted scores. It allows DNNs to output more conservative scores, which are generally are over-confident.
%Indeed model tend uniform prediction if T $\rightarrow +\infty$. 

\paragraph{Dirichlet calibration } 
Dirichlet calibration \cite{kull2019dircal} considers that score vectors follow a Dirichlet distribution. It transforms $log(F(x))$ instead of impacting the logits, considering that scores result from a softmax: $ F(x)_{c}^{cal}= \sigma\Big({W \ln \big(F(x) \big)+b}\Big)_c$, $W \in \mathbb{R}^{c\times c}$ and $b \in \mathbb{R}^{c}$ being fit on a holdout set.

% but on a log-transformed probabilities (obtained after the last softmax function) with :
% \begin{equation*}
% F(x)_{c}=\frac{e^{\left(W \ln \left(Lo_{c}\right)+b\right)}}{\sum_{j=1}^{c} e^{ (W\ln (Lo_{c})+b)} } , 
% Lo_{c}=\frac{e^{L_{c}}}{\sum_{j=1}^{C} e^{L_{j}}}
% \end{equation*}

\subsection{Interpretation Methods}
We use various methods in our experiments to interpret model's predictions. They cover a large range of approaches used for interpretation, from model-aware to model-agnostic ones.
\paragraph{Model-aware interpretation}
% \paragraph{Gradient based interpretation}
While some approaches only use gradient information like the Sensitivity method \cite{Simonyan2014DeepIC} and Guided Backpropagation \cite{Springenberg2015StrivingFS}, others combine such information with a latent representation to understand which input features led to each decision, like Grad-Cam \cite{8237336} and FullGrad \cite{srinivas2019fullgrad}. 
These gradient-based methods suffer from a few short-comings, such as neurons saturation \cite{pmlr-v70-shrikumar17a}, %  and DeconvNet \cite{10.1007/978-3-319-10590-1_53}
which have been overcome by methods averaging the gradients over a linear interpolation between the input image and a reference, like Integrated Gradient \cite{10.5555/3305890.3306024}
Some other methods avoid using gradient and directly use the predicted scores given by the model, to backpropagate it to the input of the model like DeepLIFT \cite{pmlr-v70-shrikumar17a}, a fast algorithm approximating Shapley values. 
%like LRP \cite{2015PLoSO} and DeepLIFT \cite{pmlr-v70-shrikumar17a}, the latter being a fast approximation of the Shapley values hence the method being regarded as based on local approximation of the model. 
Finally, other methods rank the importance of each latent representation like Score-Cam \cite{9150840}, which relies on a model-aware occlusion based approach.
\paragraph{Model-agnostic interpretation}
We also use methods that consider interpreted models as black-boxes, such as occlusion-based approaches which degrade the input and analyse predicted score variations to define the importance of each part of the input, like Deletion \cite{Chang2019ExplainingIC,explains}, MP~\cite{8237633} or RISE~\cite{Petsiuk2018RISERI}. 
Others methods use surrogate models in order to locally emulate the complex model behaviour in an interpretable fashion, such as LIME \cite{ribeiro2016why} and SHAP \cite{NIPS2017_8a20a862}.\\

% Deletion ref removed : Zintgraf2017VisualizingDN

We selected a subset of these interpretation methods.
To represent model-agnostic methods, we used RISE and MP, which do not rely on any other information than the output scores of the model for any given input, and Sensitivity, Integrated Gradient for model-aware methods. 
We also conducted early experiments with other model-aware interpretation methods that were little impacted by calibration, like Grad-CAM, LRP and DeepLIFT, which we chose not to include in this publication to focus on what can be useful to practitioners. 
Although these approaches are very popular, we did not evaluate LIME and SHAP methods as they are not totally suited for images and are computationally expensive.

% \emph{RISE}'s saliency map \cite{Petsiuk2018RISERI} is obtained by computing the sum  of randomly sampled masks weighted by the scores obtained when the model is applied to these masked image. Given a certain mask $m_{j}$ and the number of samples $N$, we have :
% \begin{equation}
% S_{x, F}\left(x_{i}\right) = \frac{1}{\mathbb{E}\left([m]\right)*N}\sum_{j=1}^{N}F(I\odot m_{j})(x_{i})
% \end{equation}

% \emph{Meaningful Perturbation} \cite{8237633} (MP) tries to find the smallest mask which makes the prediction score drop the most by optimizing :

% \begin{equation}
% \min _{m \in[0,1]^{H*W}} \lambda\|1-m\|_{1} + \beta TV(m) +F(x\odot m)_{c}
% \end{equation}

% with $TV$ being the total variation \cite{RUDIN1992259} of the mask. 
% This parameter penalizes the shape of the mask to be as regular as possible in order to avoid adversarial artifacts.

Evaluating the validity and limitations of these methods has been the aim of several works \cite{Kindermans2019,Adebayo_2019}, for example by measuring their sensitivity to adversarial effects \cite{Ghorbani_Abid_Zou_2019}, their alignment with human perception \cite{Mohseni_2021}, their faithfulness to the model being explained \cite{Hooker_2019}, or their stability \cite{AlvarezMelis2018OnTR}. 
We can rely on these works to build our experimental assessment of the calibration's impact on interpretations.

\section{Evaluation of Calibration's Impact on Interpretation}
\label{sec:empirical-study}

%This section presents the various experiments that have been conducted to asses the different properties of the impact of calibration on local interpretation methods.

\subsection{Objectives}

Although local interpretation approaches differ, they all rely on the output score vector $F(x)$, which is modified by the calibration process. Our aim is to assess whether or not these modifications have an impact on interpretations, and if this potential impact is rather positive or negative.

%Local interpretability algorithms introduced previously differ in the approach, however they rely on the output score vector $F(x)$, which is modified by the post-hoc calibration process.
%We aim at evaluating if these modifications on the output scores have an impact on the interpretation methods' outcomes and if this impact is rather positive or negative.

Assessing the quality of feature importance - here provided by saliency maps - is challenging, yet we argue that a good interpretation should at least: it should be \emph{faithful} to the model it explains, meaning that removing pixels defined as salient should have an impact on the model outputs \cite{jacovi2020towards}, interpretations should be \emph{robust} \cite{AlvarezMelis2018OnTR}, so that similar inputs should lead to similar interpretations, and it should be composed of structured and smoothly-varying components \cite{smilkov2017smoothgrad}, in order to respect human expectations in terms of \emph{visual coherence}.

%Let us first precise what can be expected from a good local interpretation algorithm in the scope of image classification, i.e. in terms of prediction power attribution to each feature, or pixel. The resulting attributions are generally presented as saliency maps.
%Assessing the quality of a saliency map is a challenging task but we argue that a good interpretation should have some characteristics regarding its relation with the behaviour of the model and its understandability. 
%First, it should be \emph{faithful} to the model it tries to explain, meaning that removing pixels defined as salient should have an impact on the model outputs \cite{jacovi2020towards}.
%\todo{to be removed if the corresponding section is moved to supplemental}Then interpretations should be \emph{robust} \cite{AlvarezMelis2018OnTR}, therefore similar inputs should lead to similar interpretations.
%Finally, it should be composed of structured and smoothly-varying components \cite{smilkov2017smoothgrad}, in order to respect human expectations in terms of \emph{visual coherence}.

To evaluate the impact of calibration on interpretations, we first assess if the calibration process actually impacts the resulting saliency maps by making pair-wise comparisons between them. We then quantify the impact of these changes by evaluating how the classifier's confidence drops when progressively removing important features. 
Third we assess the visual coherence of the produced saliency, by qualifying their structure and smoothly-varying properties,
and we finally evaluate the gain in stability of interpretations when calibrating a model.

%In order to address these different properties, we structure our experimental evaluations in 4 steps. 
%First we assess if the calibration process actually impacts the resulting saliency maps by making pair wise comparison of these in Section \ref{sec:impact}. 
%Second we evaluate whether the calibration impact is positive or negative with respect to the faithfulness of the produced saliency maps by comparing uncalibrated and calibrated saliency maps with random saliency map via the analysis of their respective deletion curves area.
%Third we assess the visual coherence of the produced saliency, by qualifying their structure and smoothly-varying properties. We resort to the Total Variation on the binarized saliency to produce comparable measures.
%Finally the stability of the interpretability algorithm is evaluated on calibrated or uncalibrated models to evaluate if calibration improves the robustness of the local interpretation methods.

\begin{figure}[t]
    \centering
    \includegraphics[width=0.99\textwidth] {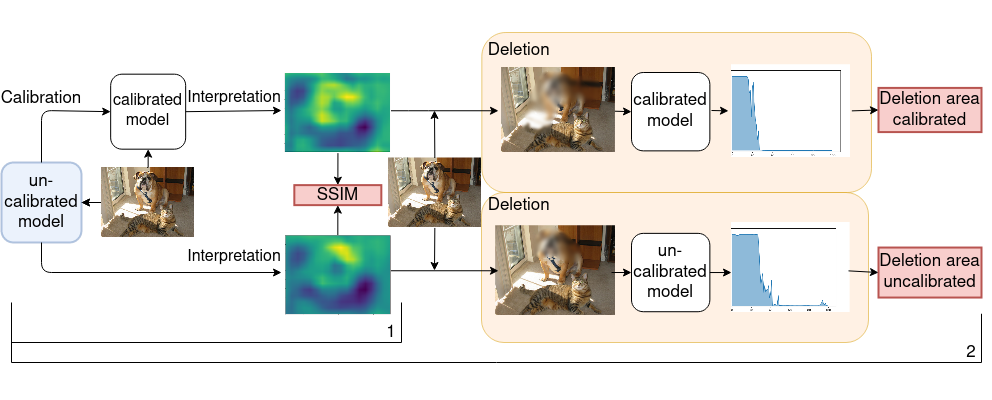}
    %scale was 0.3
    \caption{Protocols to evaluate calibration's impact on interpretation methods: \\1 - Comparison of interpretations using SSIM, and 2 - Progressive deletion impact}
    \label{fig:protocol}
\end{figure}

\subsection{Experimental Setup}
We designed cautiously an experimental setup to ensure the comparison of diverse models, image classification datasets and methods.
Our goal was to isolate carefully the impact of the calibration on the interpretation procedures.
Post-hoc calibration does not impact the models deeply, hence the observable modifications in interpretations is caused by the calibration step, as we compare the interpretation outcomes on uncalibrated models and their calibrated counterparts, all other things being equal.

\paragraph{Models}: The following experiments were conducted with VGG, RESNET and EfficientNet models, classical and diverse architectures for image classification. As DNNs are known to be overconfident \cite{pmlr-v70-guo17a,Fooled_Nguyen}, in practice most pre-trained models available in model zoos are not calibrated, and when applying standard learning algorithm we directly obtain uncalibrated models.  
Yet, to be able to observe the effect of calibration, the tasks or datasets, should be sufficiently complex for the model, so that the accuracy of the model is not perfect, so that there is no room for calibration improvements.

\paragraph{Datasets}: We chose three datasets (of various resolutions) to run our experiments : CIFAR-100\cite{krizhevsky2009learning}, Food101\cite{bossard14} and Birds (CUB-200)\cite{WahCUB_200_2011}. 
These datasets allow a proper use of calibration in its rigorous context, since each image contains exactly one instance of known classes, and they present different visual properties and classification complexities.
To ensure reproducible research, we used pretrained models on CIFAR-100 (VGG16 and RESNET50\footnote{https://github.com/chenyaofo/pytorch-cifar-models}) 
and Food-101 (ResNet50\footnote{https://github.com/Herick-Asmani/Food-101-classification-using-ResNet-50}). 
As no pre-trained EfficientNet on Birds dataset were available in open access, we fine-tuned (100 epochs with a learning rate of 1e-4 using Adam optimizer) an EfficientNet pretrained on ImageNet (from torchvision) onto the Birds dataset.

\paragraph{Calibration}: The calibration step has been performed using the calibration methods described previously (Temperature Scaling and Dirichlet calibration). For both methods and every model, we used a calibration set composed of 3000, 2500 and 2500 samples taken from the test set for CIFAR-100, Food-101 and Birds respectively. We evaluated the ECE before and after the calibration using the continuous estimator $ECE_{density}^{conf}$ introduced in \cite{posocco2021ece} using the bandwidth automatically set with Silverman's rule,  on 2500 samples.
ECEs and accuracies of the different models, given in Table~\ref{tab1}, confirm the initial mis-calibration of the raw models, and the effectiveness of the calibration step.

\paragraph{Interpretation}: For Integrated Gradients (IG), black and white references combined are used with 30 equidistant points on the convex path from the reference to the input. 
For RISE, we randomly sample 4000 8x8 binary masks (higher dimensions would require more sampling), upscale them using bicubic interpolation, and values of the mask are drawn from a $0.6$ Bernoulli law. 
For MP, optimization problems are solved using the Adam optimizer ($\alpha = 0.1, \beta = 0.4 , lr=0.
1$) for 600 steps (these optimization parameters have not been fine-tuned).\\

%For LRP, a jitter $\epsilon = 3$ is used. Note that LRP has not been run on the ResNet models as it is not compatible with residual connections. 
%Finally Sensitivity (S) do not need configuration.  
All computed saliency maps are min-max-normalized. Some of these, resulting from each of the method applied on calibrated and uncalibrated model, can be observed in Figure~\ref{fig:saliency-example}. 
The saliency maps analyzed in the following experiments were obtained from 2500 images randomly sampled from the remaining test set of each of the datasets (which have not been used for calibration).

\begin{table}[t]
\caption{Accuracy and confidence-calibration of used models}
\begin{center}
\label{tab1}
  \begin{tabular}{|c|c|c|ccc|}
  \hline
    \multirow{2}{*}{Model} &
    \multirow{2}{*}{Dataset} &
    \multirow{2}{*}{Accuracy} &
        \multicolumn{3}{c|}{Confidence calibration ($ECE^{conf}$)} \\
    & &  &
Base & Temperature & Dirichlet \\
    \hline
    \multirow{2}{*}{VGG16} & Food101 & 0.4470 & 0.1997 & 0.03 & 0.063 \\
    & Cifar100 & 0.6811 & 0.2003 & 0.0442 & 0.0477 \\
    \hline
    \multirow{1}{*}{RESNET32} & Cifar100 & 0.6371  & 0.1484 & 0.0463 & 0.0323 \\
    \hline
    \multirow{1}{*}{RESNET50} & Food101 & 0.8173 & 0.0803 & 0.0323 & 0.0463 \\
    \hline
    \multirow{1}{*}{EFFICIENTNETB0} & Birds & 0.7984 & 0.0706 & 0.0234 & 0.0093 \\
    \hline
\end{tabular}
\end{center}
\end{table}

\subsection{Does Calibration Impact Interpretations ?}
\label{sec:impact}

\begin{figure*}[tbh]
    \includegraphics[width=\textwidth]{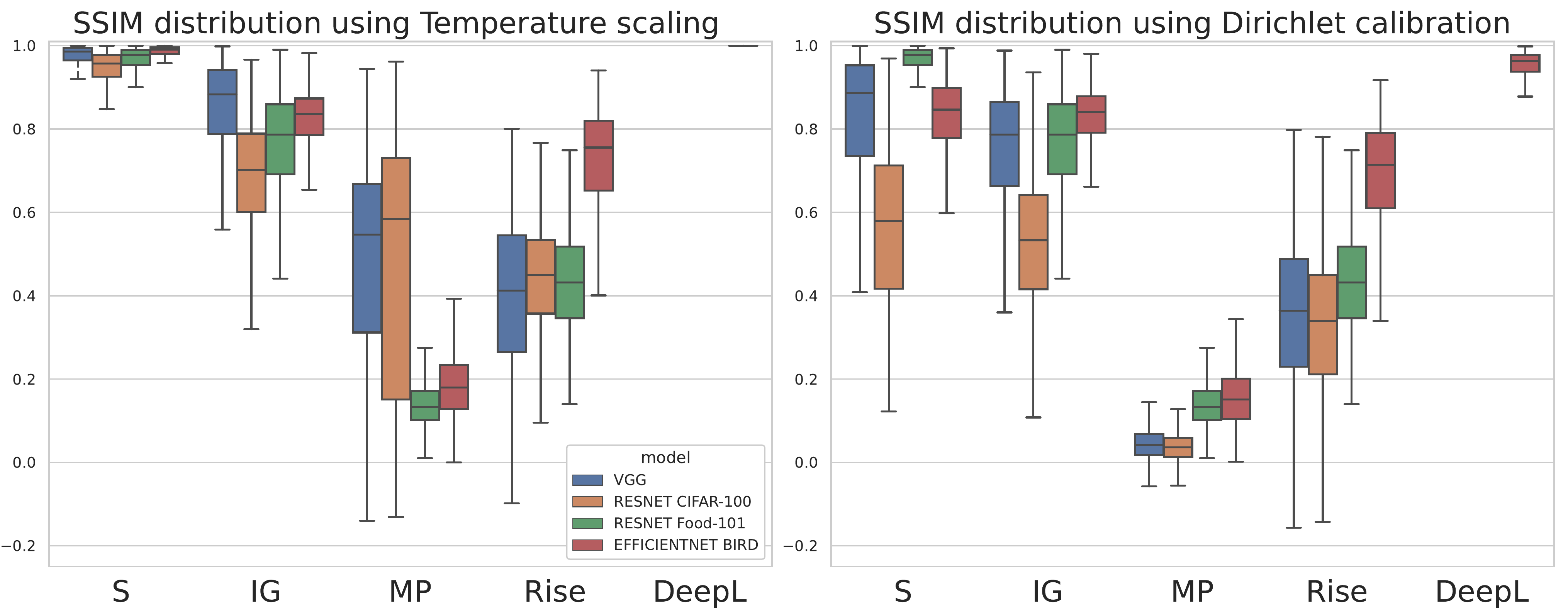}
    %\centering
    \caption{Distribution of the SSIM between interpretations from calibrated and uncalibrated models}
    \label{fig:SSIM-res}
\end{figure*}

We start by comparing each pair of interpretations - coming from uncalibrated and calibrated models - using the structural similarity index (SSIM \cite{Wang04imagequality}), which is based on human perception and thus more relevant than MAE or MSE. This experimental protocol is synthesised in Figure~\ref{fig:protocol} part 1.

The results of this experiment, presented in Figure \ref{fig:SSIM-res}, show that for all studied interpretation methods, there is a significant difference between saliency maps produced from the calibrated and uncalibrated models.
% It is consistent with the fact that MP and Rise are model-agnostic, and use the scores directly, contrary to Sensitivity and Integrated Gradient.
This impact is particularly important on model-agnostic interpretation methods like MP and RISE.

\subsection{Does Calibration Improve the Faithfulness of Interpretation Methods ?}
\label{sec:faithfulness}

We further investigate this impact to determine if detected visual changes in saliency maps improve their quality in terms of \emph{faithfulness}~\cite{jacovi2020towards}.
For each pair of interpretations built by the previous experiment, we apply a deletion procedure \cite{Petsiuk2018RISERI}: input features are ranked according to their importance given by the interpretation, and are progressively neutralized while we observe the impact on the score of the predicted class from the degraded input image. 
% At each percentage of removed pixels, the models make a prediction on the damaged input in order to model by a curve the score drop with respect to the percentage of removed pixels. 
We compute the \emph{deletion area} (area under the score curve wrt to the percentage of neutralized pixels). 
In practice, to preserve the input's distribution, neutralized pixels are replaced by an 11x11 gaussian blurred patch ($\sigma=10$) centered on this pixel. 
The whole procedure is summed up in Figure~\ref{fig:protocol} part 2. 

Additionally to the four interpretation methods we use a random interpretation baseline as reference for the comparison.
We remove a given percentage of ~100 randomly sorted superpixels computed with the SLIC algorithm \cite{6205760}.
% To use these deletion area in practice, we compare them with a reference baseline: we first compute superpixels of the input image using the SLIC algorithm \cite{6205760} starting with 100 centroïds --the goal is to work at the level of meaningful regions instead of independent pixels--, we then randomly masked these superpixels and computed the deletion area. The convergence of SLIC does not guarantee that we end up with 100 superpixels, so to be able to compare the baseline deletion curves are linearly interpolated to 100 points. 
We averaged the deletion curve obtained with five different random orders.

We  compute the  deletion area on the random baseline, uncalibrated models and calibrated ones. 
Uncalibrated models are more confident, hence to ensure a fair comparison, we normalize the deletion curves in order to set the score of the predicted class from the initial clean image to one. 
The normalized curve then shows how much the confidence of the model drops, with respect to its initial confidence, when we neutralize pixels in decreasing importance.\\ 

The different interpretation methods show consistent results over the various datasets. Before any calibration considerations, they show very different level of faithfulness, Meaningfull perturbation being the most faithful with a predicted score quickly dropping as the percentage of neutralized pixels grows, followed by RISE, the other model-agnostic method.  
MP and RISE are also the most positively impacted by the calibration of the model. 
These two observations are confirmed by the measurement of the deletion areas shown in the second line of the Figure \ref{fig:general_comparison_graph}.
For the other interpretation methods, the calibration procedure shows little impact, as expected from the SSIM experiment. 
% Figure \ref{deletion_curves} shows that, on CIFAR-100, the mean of the deletions curves obtained on the entire dataset is not modified by the calibration process for Sensitivity, LRP and Integrated Gradients and Rise. This means that the changes induced by calibration, on average, neither have a positive nor a negative effect on the faithfulness of these method. However, MP benefits from a real improvement when used on a calibrated model. This is interesting as one can also observe that MP method have the best performances as evaluated by the deletion curve.
% On Food-101, there is a noticeable improvement for both Rise and MP.
% We hypothesize that calibration increase Rise known better capabilities in higher dimensions leading to this different behaviour between CIFAR-100 and Food-101.
One can also confirm a posteriori that the normalization of the deletion curves does not introduce any bias, since the same random saliency maps applied to the calibrated and uncalibrated models produce similar curves.

We conduct another analysis to assess the gain that calibration brings in an element-wise comparison of the deletion area. 
We compare each interpretation with the random baseline and consider that an prediction is well explained if the deletion area is lower for the method than the one obtained with the random saliency. The proportion of well explained images are reported in Figure~\ref{fig:general_comparison_graph} (third line) as Better Than Random ratio (BTR).
This BTR is always improved, except for one model/method case. The improvement varies from limited to important depending on the approach and the dataset.
We read the greater impact of the Dirichlet calibration wrt the Temperature scaling, on both the BTR and deletion area, besides that their respective ECE values are cases comparable in most cases, by the fact that Dirichlet calibration can change the predicted class while temperature scaling does not.

% There is no notable change on the BTR for Temperature Scaling, which means calibration does not improve the ability to interpret predictions that already could not be interpreted before. However there is an improvement coming from Dirichlet calibration, which may come from the fact that it can change the predicted class, therefore the gain is not more related to the intrinsic abilities of the interpretation methods but come from the relation between the model and it's calibration.

The great improvement brought by calibration to MP is promising for model-agnostic interpretation. Notably, it is known that without computationally expensive hyperparameters tuning, MP is sensitive to visual artifacts \cite{article}.
\begin{figure}[t]
    \centering
    \includegraphics[scale=0.25]{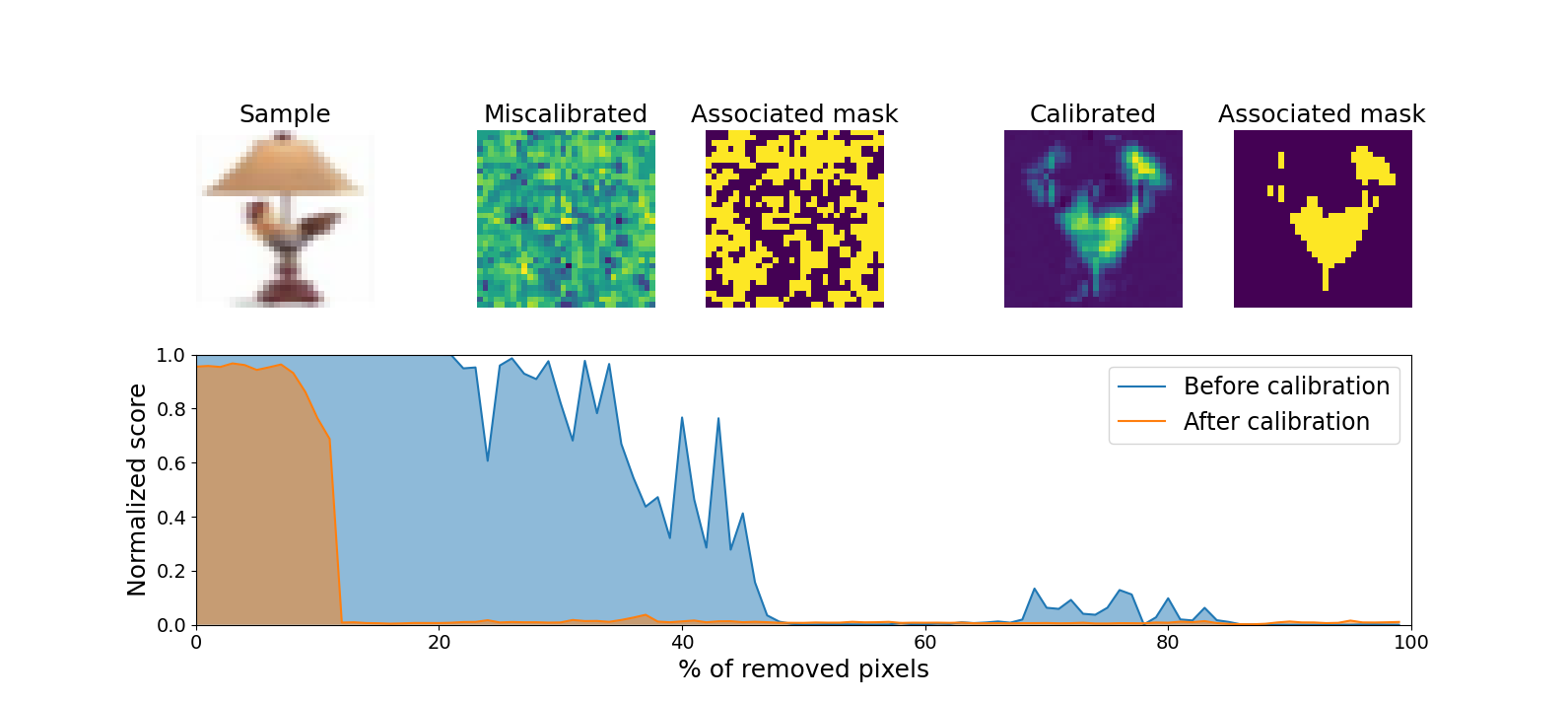}
    \caption{Saliency maps, Otsu-binarized masks and deletion curves, for the calibrated model and the uncalibrated one, using MP, on a given sample. Map explaining the calibrated model is consistent with human expectation and more faithful to the model.}
    \label{artifacts}
\end{figure}  
As the example in Figure \ref{artifacts} suggests, calibration seems to help dealing with those.
%The neat advantage of MP with regard to the faithfulness evaluation comes at a price of an higher computational cost.

% the in the worst case there is no negative impact 
To sum up, we measure a positive impact on the faithfulness as measured by the deletion area and the BTR ratio presented in Section~\ref{sec:faithfulness} for model-agnostic methods, in worst case the interpretability is not impacted by the calibration.
The most faithful method without calibration, namely MP, is also the most improved method. 
This suggests that the interpretability of the model in itself could depend on its calibration.

% Figure \ref{fig:mean_auc} exhibits that the minimum of the deletion curves are obtained when the model is calibrated, showing a clear non linear correlation between the temperature scaler value and the deletion area.
% 
%\begin{figure}[h]
%    \centering
%    \includegraphics[scale=0.25]{mp_auc.eps}
%    \caption
%    	{Mean deletion area on CIFAR-100 for VGG16 and RESNET32 conditioned on the temperature inverse. Marked points indicate where the temperature for which the model is calibrated.}
%    	\label{fig:mean_auc}
%\end{figure}

%aucs_vgg_temp.eps
\begin{figure}[tbh!]
    \centering
    \includegraphics[width=1\textwidth]{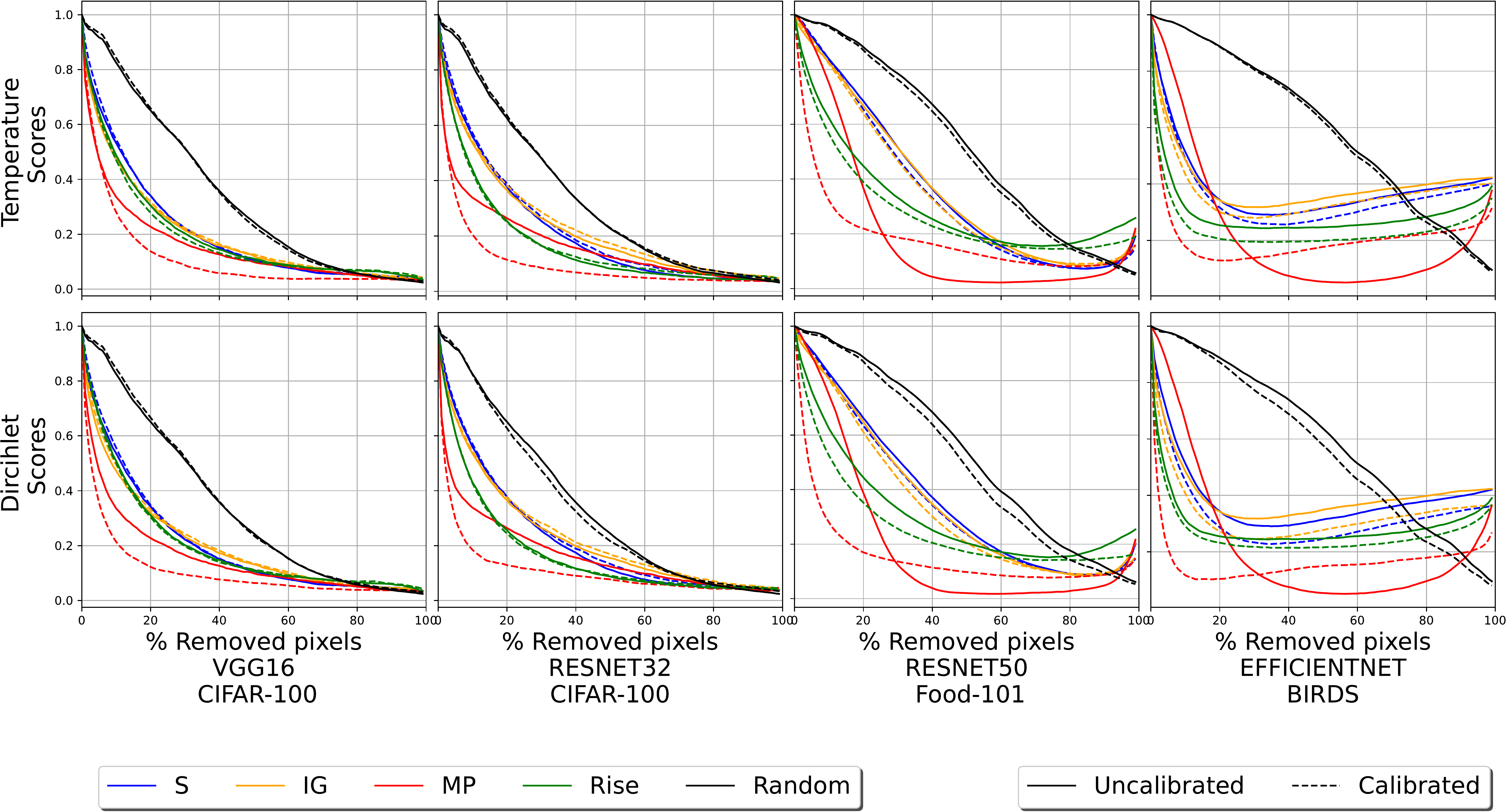}
    \caption{Deletion curves before (plain line) and after calibration (dashed line) for three different models using Temperature Scaling (upper) and Dirichlet calibration (lower)}
    \label{deletion_curves}
\end{figure}

% \subsection{Homogeneity of produced saliency maps}
\subsection{Are Saliency Maps with Calibration more Human-Friendly ?}
\label{sec:TV}

Another important aspect of  interpretation is the readability for users, smoothly varying saliency maps are easier to comprehend,
% We cannot evaluate the relevance of an interpretation without analyzing its complexity. Sparser interpretations with continuous activated zones are usually considered as more understandable,
as they tend to highlight structures that we, as humans, recognize.
Therefore, to quantify the complexity of a produced saliency map, we first distinguish activated pixels (foreground of saliency maps) from non-activated ones (background) using Otsu binarization \cite{4310076}, and compute the total variation (TV) of obtained binary images. A higher total variation suggests a more noisy interpretation. 

\begin{figure*}[t]
    \centering
    \includegraphics[width=0.99\textwidth]{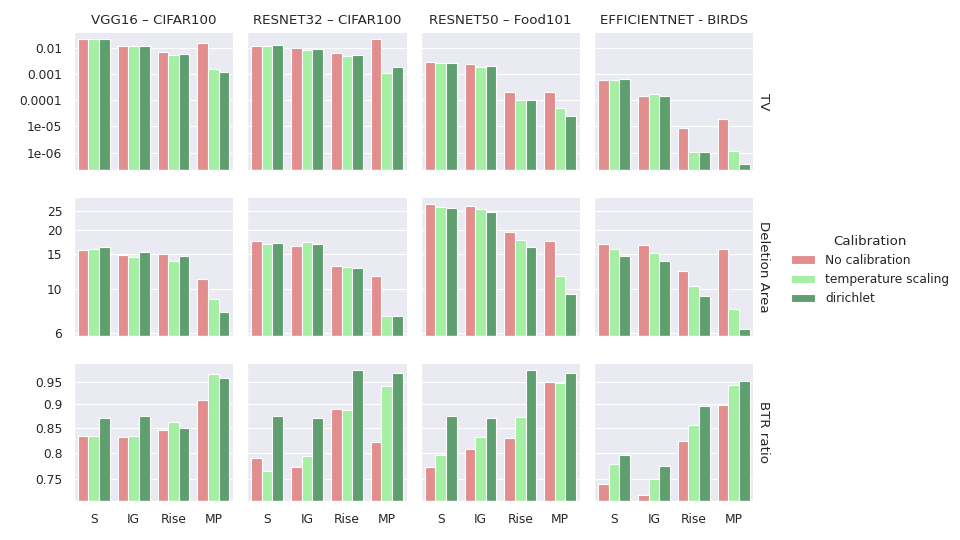}
    \caption{Ratio of well explained images (BTR Ratio), Otsu-binarized TV and Deletion Area obtained by various interpretation methods on multiple models and datasets before and after calibration.}
    \label{fig:general_comparison_graph}
\end{figure*}

Figure~\ref{fig:general_comparison_graph} shows that Otsu-TV is always improved (lowered) by calibration for model-agnostic methods, or at worst unaffected for other methods,
% that LRP produces the clearest saliency maps. 
% Indeed, it is highly sensitive to high frequencies and % % properly higlights the edges. 
% LRP's total variation is not modified by calibration. 
which means that interpretations exhibit smoother variations and fewer highlighted regions, while the mean deletion area is constant or improved. Hence the interpretations of the calibrated models are more human readable while being equally or more faithful to the model.

%\subsection{Additional samples for visualising calibration impact}
%Additionally to the figure presented in introduction, we show in Figure~\ref{fig:wall} some samples with corresponding saliency maps computed from various methods on a VGG16 model trained on the CIFAR-100 dataset, with and without calibration via temperature scaling.
%\begin{figure}[h]
    %\begin{adjustwidth}{-6.35cm}{2cm}
    %\includegraphics[scale=0.5]{wall.eps}
    %\end{adjustwidth}
    %\caption
    	%{Some saliency maps samples on VGG16 for CIFAR-100 calibrated or not with temperate scaling}\label{fig:wall}
%\end{figure} 

\subsection{In Depth Analysis of Meaningful Perturbation}
\label{sec:robust}
To better understand our findings, we focus now on MP, the most faithful method, which appears to be the most impacted by the calibration.
%Let us now focus on MP to highlight reasons for our empirical observations.

\begin{figure}[t]
    \centering
    \includegraphics[width=\textwidth]{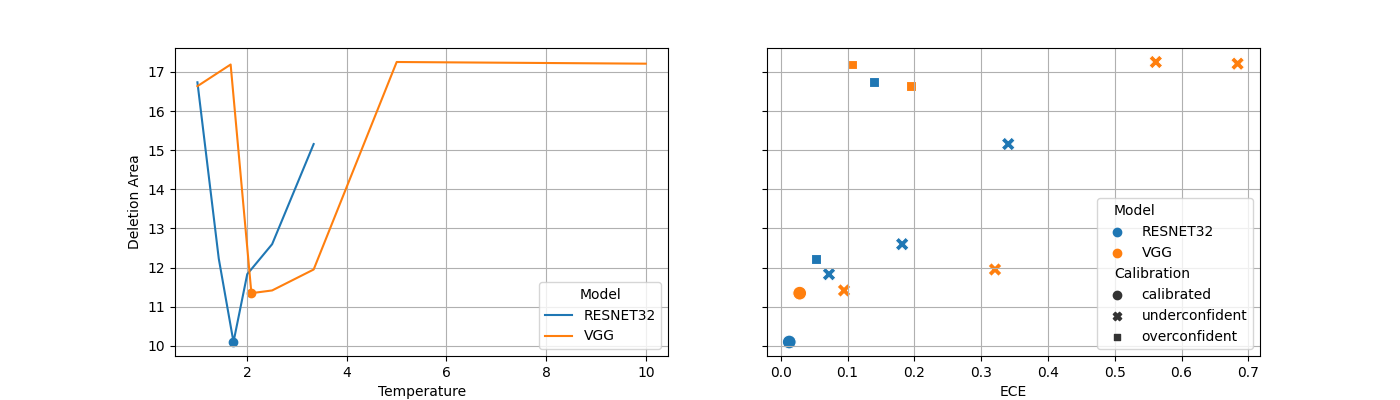}
    % \subfigure{\includegraphics[scale=0.27]{inv_temp.png}}
    % \subfigure{\includegraphics[scale=0.30]{ece_deletion.eps}}
    % \subfigure{\includegraphics[scale=0.30]{mean_auc_last_point.png}}
    \caption
    	{Mean deletion area on CIFAR-100 for VGG16 and RESNET32 conditioned on the temperature. Marked points indicate calibrated models.}\label{auc_plots}
\end{figure}

\paragraph{Effect of the mis-calibration}
In order to evaluate if, for MP, the faithfulness improvement correlates with the mis-calibration level, we apply the deletion experiment using Temperature Scaling for different fixed temperatures. 
Figure \ref{auc_plots} highlights that the minimum of the deletion area is obtained when the model is calibrated, showing a clear non linear correlation between the scaler's temperature value and the deletion area. 
Interestingly the second plot highlights a positive correlation between the mean AUC (deletion area) and the ECE. The correlation differs whether the mis-calibration is due to an overconfident model or an underconfident model for the same range of ECE. The overconfident models are those with a temperature smaller than the best obtained. 

% Mean deletion area on CIFAR-100 for VGG16 and RESNET32 conditioned to the inverse of the temperature (left) and on the expected calibration error (right). Marked points indicate the temperature for which the model is calibrated (left). On the right black triangles show when the model is underconfident ($T > T_{opt}$) and red squares when it is overconfident ($1\leq T < T_{opt}$).
 
\paragraph{Interpretation stability}
As shown previously, calibration improves the deletion area and the total variation of saliency maps from MP, the best method in terms of deletion area, meaning that produced interpretations are both more faithful (according to \cite{jacovi2020towards} definition) and more spatially coherent.
One possible explanation for this improvement is that calibration improves the stability of the method. It is known that MP is sensitive to visual artifacts \cite{article} without proper hyperparameters tuning. 
Calibration seems to be a really adequate solution to prevent the apparition of such artifacts in such artifacts, especially when parameter tuning is not feasible. 
Indeed, as underlined in \cite{AlvarezMelis2018OnTR}, interpretation methods tend to be unstable when the input is slightly modified. 
As calibration is related to the model's robustness \cite{robutness} it is natural to wonder whether or not the calibration of a given model improves the robustness of the interpretations made on its predictions. 
To that end, we compute an approximation of the Lipschitz constant of the saliency function, following the experiment introduced in \cite{AlvarezMelis2018OnTR}, by adding small amount of gaussian noise to the input, so that the behaviour of the model does not change, and compute the normalized $l_2$-norm of the interpretations obtained with and without calibration.
This experiment was applied on 150 randomly sampled points from CIFAR-100. For each point, we sample 40 neighbour points (inside the ball of radius 0.05 and with the point of interest as center).
We only apply this experiments on 150 points in order to keep it feasible in a reasonable amount of time (for MP and Rise processing each point takes around an hour with an NVIDIA 1650).
\begin{figure}[t]
    \centering
    \includegraphics[width=\textwidth] {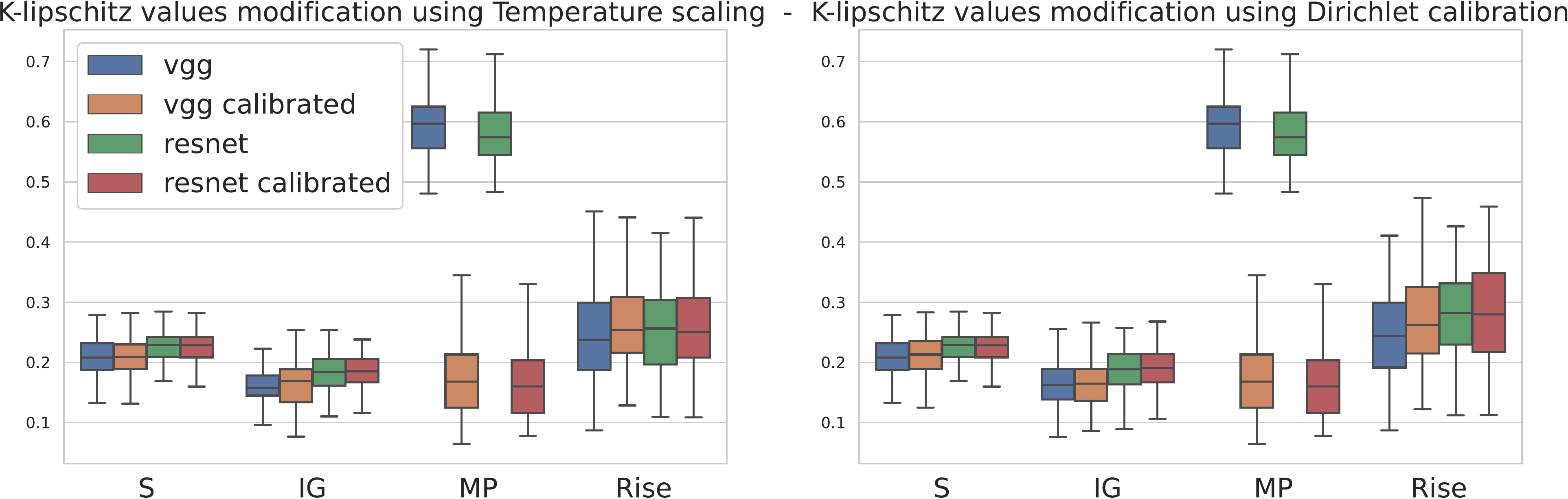}
    \caption{K-lipschitz distribution obtained with calibrated and uncalibrated model using Temperature Scaling (left) and Dirichlet calibration (right).}
    \label{Klipschitz}
\end{figure}

Figure \ref{Klipschitz} reveals that calibration, while having no robustness impact on most methods, considerably improves the stability of MP. 
This is consistent with the  $10\times$ decrease of TV for MP revealed in the previous section. % is such that it is probably the results of an improvement in the robustness of the interpretation method. 
Also, RISE, which is the least stable even after calibration, could be improved by increasing the number of sampled masks at the expense to an higher computational cost.

\subsection{Discussions}
While these results shed light on non-trivial interaction between interpretation methods and calibration, it is hard to express from the gain in term of deletion area a definitive statement about interpretation methods validity. 
This come from the difficulty of evaluating explanation and therefore to the best of our knowledge no consensus has been attained in terms of the best method to validate interpretation methods.
% and therefore each evaluation process of interpretability methods has its drawbacks. 
The deletion experiments, while truly evaluating interpretations quality, seems to favor methods which highlight regions over methods which highlight high frequency details, which would account for the poor faithfulness performance as measured by the deletion area. 
Moreover, we are convinced that a good interpretation method should be able to reflect the uncertainty of the model and therefore should be impacted by the calibration, we even argue that this could be a clue to setup a new sanity check for interpretation methods.

\section{Conclusions and Future works}
\label{sec:concl}

This paper studies the relationship between uncertainty and explainability, two important aspects of trustworthy machine learning. 
More specifically we evaluate the impact of post-hoc calibration of a given image classification model over the quality of the saliency maps produced by several widely used local interpretation methods applied on the model for different datasets and models.\todo{add that these have been chossed because they are plug and play}
The experimental benchmark, built to evaluate this impact \emph{all other things being equal}, shows a positive impact of calibration on produced interpretations, in terms of faithfulness, stability and visual coherence. 
% Calibration and interpretability do have an impact on each other for some widely used approaches. 
The impact, particularly beneficial on model-agnostic interpretation methods such as Meaning Perturbation (MP)~\cite{8237633}, is in the worst case neutral.
For these reasons, we suggest a simple practice to improve interpretation outcomes : \emph{Calibrate to Interpret}.

A side benefit of the study is to rank the competing interpretation approaches, which shows that model-agnostic ones perform better with regard to faithfulness, measured by the deletion area, and visual coherence, measured by the Otsu-TV.
Interestingly the stronger impact of calibration appears on the best interpretation method namely MP, and resolves one of its main drawbacks: its sensitivity to artifacts. 
We highlight for this method that there is a clear correlation between the calibration level of a model and the faithfulness of the MP method applied to its predictions.
% Moreover we show an interesting positive correlation between the level of calibration of a model and the faithfulness of the MP method applied on it.

%With recent applications of DNN to critical contexts, there is an urge to develop solutions that would be able to qualify and even certify a set of model properties, such as the fairness, interpretability, robustness or uncertainty, in order to help end users trusting these deep models. 
%This work contributes to connect these properties, by showing that interpretations and uncertainty are dependent and that it is worthwhile to calibrate a model, not only to have relevant probability scores, but also to have faithful and cleaner local interpretations. 

This work opens the road to deeper analyses.
A direct extension would be to analyse if \emph{in-training} solutions to enforce calibration would impact similarly the interpretations. This would require a different benchmark setup, where modifications induced to the model by \emph{in-training} calibration are properly framed.
% For example, the improvement of the TV for RISE and MP indicates that the interpretations obtained are more spatially coherent. It would be interesting to evaluate if it correctly identifies class instances on multilabel classification. 
% Then, it would be interesting to evaluate if the interpretations of the calibrated model are preferred over the miscalibrated one by an end user.
Additionally, although computational cost considerations prevented us from realizing experiments with the ROAR procedure~\cite{hooker2018evaluating} to measure the faithfulness of interpretation methods, we think the theoretical properties of this approach make it a great option to deepen our experimental evaluation.  
Finally, we wonder if other kinds of explanation paradigm, e.g. concept based~\cite{chen2019looks}, sample based~\cite{pruthi2020estimating} or even attention based~\cite{chen2020generative} can also benefit from calibration.

\bibliographystyle{splncs04}

\begin{thebibliography}{10}
\providecommand{\url}[1]{\texttt{#1}}
\providecommand{\urlprefix}{URL }
\providecommand{\doi}[1]{https://doi.org/#1}

\bibitem{6205760}
Achanta, R., Shaji, A., Smith, K., Lucchi, A., Fua, P., Süsstrunk, S.: Slic
  superpixels compared to state-of-the-art superpixel methods. IEEE
  Transactions on Pattern Analysis and Machine Intelligence  \textbf{34}(11),
  2274--2282 (2012). \doi{10.1109/TPAMI.2012.120}

\bibitem{Adebayo_2019}
Adebayo, J., Gilmer, J., Muelly, M., Goodfellow, I., Hardt, M., Kim, B.: Sanity
  checks for saliency maps. In: Advances in Neural Information Processing
  Systems. p. 9525–9536. NIPS'18, Curran Associates Inc., Red Hook, NY, USA
  (2018)

\bibitem{AlvarezMelis2018OnTR}
Alvarez-Melis, D., Jaakkola, T.S.: On the robustness of interpretability
  methods. arXiv preprint arXiv:1806.08049  (2018)

\bibitem{bossard14}
Bossard, L., Guillaumin, M., Van~Gool, L.: Food-101 -- mining discriminative
  components with random forests. In: European Conference on Computer Vision
  (2014)

\bibitem{Chang2019ExplainingIC}
Chang, C.H., Creager, E., Goldenberg, A., Duvenaud, D.: Explaining image
  classifiers by counterfactual generation. In: International Conference on
  Learning Representations (2019)

\bibitem{chen2019looks}
Chen, C., Li, O., Tao, C., Barnett, A.J., Su, J., Rudin, C.: This looks like
  that: deep learning for interpretable image recognition. In: Advances in
  Neural Information Processing Systems. p. 8928–8939 (2019)

\bibitem{chen2020generative}
Chen, M., Radford, A., Child, R., Wu, J., Jun, H., Luan, D., Sutskever, I.:
  Generative pretraining from pixels. In: International Conference on Machine
  Learning. pp. 1691--1703. PMLR (2020)

\bibitem{article}
Fong, R., Vedaldi, A.: Interpretable explanations of black boxes by meaningful
  perturbation. pp. 3449--3457 (10 2017). \doi{10.1109/ICCV.2017.371}

\bibitem{8237633}
Fong, R.C., Vedaldi, A.: Interpretable explanations of black boxes by
  meaningful perturbation. In: 2017 IEEE International Conference on Computer
  Vision (ICCV). pp. 3449--3457 (2017). \doi{10.1109/ICCV.2017.371}

\bibitem{Ghorbani_Abid_Zou_2019}
Ghorbani, A., Abid, A., Zou, J.: Interpretation of neural networks is fragile.
  Proceedings of the AAAI Conference on Artificial Intelligence
  \textbf{33}(01),  3681--3688 (Jul 2019). \doi{10.1609/aaai.v33i01.33013681},
  \url{https://ojs.aaai.org/index.php/AAAI/article/view/4252}

\bibitem{pmlr-v70-guo17a}
Guo, C., Pleiss, G., Sun, Y., Weinberger, K.Q.: On calibration of modern neural
  networks. In: Precup, D., Teh, Y.W. (eds.) Proceedings of the 34th
  International Conference on Machine Learning. Proceedings of Machine Learning
  Research, vol.~70, pp. 1321--1330. PMLR (06--11 Aug 2017),
  \url{http://proceedings.mlr.press/v70/guo17a.html}

\bibitem{hooker2018evaluating}
Hooker, S., Erhan, D., Kindermans, P.J., Kim, B.: Evaluating feature importance
  estimates  (2018)

\bibitem{Hooker_2019}
Hooker, S., Erhan, D., Kindermans, P.J., Kim, B.: A benchmark for
  interpretability methods in deep neural networks. In: Wallach, H.,
  Larochelle, H., Beygelzimer, A., d\textquotesingle Alch\'{e}-Buc, F., Fox,
  E., Garnett, R. (eds.) Advances in Neural Information Processing Systems 32,
  pp. 9737--9748. Curran Associates, Inc. (2019),
  \url{http://papers.nips.cc/paper/9167-a-benchmark-for-interpretability-methods-in-deep-neural-networks.pdf}

\bibitem{jacovi2020towards}
Jacovi, A., Goldberg, Y.: Towards faithfully interpretable nlp systems: How
  should we define and evaluate faithfulness? arXiv preprint arXiv:2004.03685
  (2020)

\bibitem{kim2020puzzle}
Kim, J.H., Choo, W., Song, H.O.: Puzzle mix: Exploiting saliency and local
  statistics for optimal mixup. In: International Conference on Machine
  Learning. pp. 5275--5285. PMLR (2020)

\bibitem{Kindermans2019}
Kindermans, P.J., Hooker, S., Adebayo, J., Alber, M., Sch{\"u}tt, K.T.,
  D{\"a}hne, S., Erhan, D., Kim, B.: The (Un)reliability of Saliency Methods,
  pp. 267--280. Springer International Publishing, Cham (2019).
  \doi{10.1007/978-3-030-28954-6_14},
  \url{https://doi.org/10.1007/978-3-030-28954-6\_14}

\bibitem{explains}
Kononenko, I., Robnik-Sikonja, M.: Explaining classifications for individual
  instances. IEEE Transactions on Knowledge \& Data Engineering
  \textbf{20}(05),  589--600 (may 2008). \doi{10.1109/TKDE.2007.190734}

\bibitem{krizhevsky2009learning}
Krizhevsky, A., Hinton, G., et~al.: Learning multiple layers of features from
  tiny images  (2009)

\bibitem{Beta_Kull2017}
Kull, M., Filho, T.S., Flach, P.: {Beta calibration: a well-founded and easily
  implemented improvement on logistic calibration for binary classifiers}. In:
  Singh, A., Zhu, J. (eds.) Proceedings of the 20th International Conference on
  Artificial Intelligence and Statistics. Proceedings of Machine Learning
  Research, vol.~54, pp. 623--631. PMLR (20--22 Apr 2017),
  \url{https://proceedings.mlr.press/v54/kull17a.html}

\bibitem{kull2019dircal}
Kull, M., Nieto, M.P., K{\"a}ngsepp, M., Silva~Filho, T., Song, H., Flach, P.:
  Beyond temperature scaling: Obtaining well-calibrated multi-class
  probabilities with dirichlet calibration. In: Advances in Neural Information
  Processing Systems. pp. 12295--12305 (2019)

\bibitem{pmlr-v80-kumar18a}
Kumar, A., Sarawagi, S., Jain, U.: Trainable calibration measures for neural
  networks from kernel mean embeddings. In: Dy, J., Krause, A. (eds.)
  Proceedings of the 35th International Conference on Machine Learning.
  Proceedings of Machine Learning Research, vol.~80, pp. 2805--2814. PMLR
  (10--15 Jul 2018), \url{http://proceedings.mlr.press/v80/kumar18a.html}

\bibitem{NIPS2017_8a20a862}
Lundberg, S.M., Lee, S.I.: A unified approach to interpreting model
  predictions. In: Guyon, I., Luxburg, U.V., Bengio, S., Wallach, H., Fergus,
  R., Vishwanathan, S., Garnett, R. (eds.) Advances in Neural Information
  Processing Systems. vol.~30. Curran Associates, Inc. (2017)

\bibitem{Mohseni_2021}
Mohseni, S., Block, J.E., Ragan, E.: Quantitative evaluation of machine
  learning explanations: A human-grounded benchmark. In: 26th International
  Conference on Intelligent User Interfaces. p. 22–31. IUI '21, Association
  for Computing Machinery, New York, NY, USA (2021).
  \doi{10.1145/3397481.3450689}, \url{https://doi.org/10.1145/3397481.3450689}

\bibitem{Fooled_Nguyen}
Nguyen, A., Yosinski, J., Clune, J.: Deep neural networks are easily fooled:
  High confidence predictions for unrecognizable images. In: 2015 IEEE
  Conference on Computer Vision and Pattern Recognition (CVPR). pp. 427--436
  (2015). \doi{10.1109/CVPR.2015.7298640}

\bibitem{4310076}
Otsu, N.: A threshold selection method from gray-level histograms. IEEE
  Transactions on Systems, Man, and Cybernetics  \textbf{9}(1),  62--66 (1979).
  \doi{10.1109/TSMC.1979.4310076}

\bibitem{Pakdaman_Naeini_Cooper_Hauskrecht_2015}
Pakdaman~Naeini, M., Cooper, G., Hauskrecht, M.: Obtaining well calibrated
  probabilities using bayesian binning. Proceedings of the AAAI Conference on
  Artificial Intelligence  \textbf{29}(1) (Feb 2015),
  \url{https://ojs.aaai.org/index.php/AAAI/article/view/9602}

\bibitem{Petsiuk2018RISERI}
Petsiuk, V., Das, A., Saenko, K.: Rise: Randomized input sampling for
  explanation of black-box models. In: BMVC (2018)

\bibitem{Platt99probabilisticoutputs}
Platt, J.C.: Probabilistic outputs for support vector machines and comparisons
  to regularized likelihood methods. In: ADVANCES IN LARGE MARGIN CLASSIFIERS.
  pp. 61--74. MIT Press (1999)

\bibitem{pleiss2017fairness}
Pleiss, G., Raghavan, M., Wu, F., Kleinberg, J., Weinberger, K.Q.: On fairness
  and calibration. Advances in neural information processing systems
  \textbf{30} (2017)

\bibitem{posocco2021ece}
Posocco, N., Bonnefoy, A.: Estimating expected calibration errors. In:
  Farka{\v{s}}, I., Masulli, P., Otte, S., Wermter, S. (eds.) Artificial Neural
  Networks and Machine Learning -- ICANN 2021. pp. 139--150. Springer
  International Publishing, Cham (2021)

\bibitem{pruthi2020estimating}
Pruthi, G., Liu, F., Kale, S., Sundararajan, M.: Estimating training data
  influence by tracing gradient descent. In: Larochelle, H., Ranzato, M.,
  Hadsell, R., Balcan, M.F., Lin, H. (eds.) Advances in Neural Information
  Processing Systems. vol.~33, pp. 19920--19930. Curran Associates, Inc.
  (2020),
  \url{https://proceedings.neurips.cc/paper/2020/file/e6385d39ec9394f2f3a354d9d2b88eec-Paper.pdf}

\bibitem{robutness}
Qin, Y., Wang, X., Beutel, A., Chi, E.: Improving uncertainty estimates through
  the relationship with adversarial robustness  (06 2020)

\bibitem{ribeiro2016why}
Ribeiro, M.T., Singh, S., Guestrin, C.: "why should i trust you?": Explaining
  the predictions of any classifier. In: Proceedings of the 22nd ACM SIGKDD
  International Conference on Knowledge Discovery and Data Mining. p.
  1135–1144. KDD '16, Association for Computing Machinery (2016).
  \doi{10.1145/2939672.2939778}

\bibitem{Robtuness_Interp_Slavin}
Ross, A.S., Doshi-Velez, F.: Improving the adversarial robustness and
  interpretability of deep neural networks by regularizing their input
  gradients. In: AAAI conference on artificial intelligence (2018)

\bibitem{8237336}
Selvaraju, R.R., Cogswell, M., Das, A., Vedantam, R., Parikh, D., Batra, D.:
  Grad-cam: Visual explanations from deep networks via gradient-based
  localization. In: 2017 IEEE International Conference on Computer Vision
  (ICCV). pp. 618--626 (2017). \doi{10.1109/ICCV.2017.74}

\bibitem{pmlr-v70-shrikumar17a}
Shrikumar, A., Greenside, P., Kundaje, A.: Learning important features through
  propagating activation differences. In: Precup, D., Teh, Y.W. (eds.)
  Proceedings of the 34th International Conference on Machine Learning.
  Proceedings of Machine Learning Research, vol.~70, pp. 3145--3153. PMLR
  (06--11 Aug 2017), \url{http://proceedings.mlr.press/v70/shrikumar17a.html}

\bibitem{Simonyan2014DeepIC}
Simonyan, K., Vedaldi, A., Zisserman, A.: Deep inside convolutional networks:
  Visualising image classification models and saliency maps. CoRR
  \textbf{abs/1312.6034} (2014)

\bibitem{smilkov2017smoothgrad}
Smilkov, D., Thorat, N., Kim, B., Vi{\'e}gas, F., Wattenberg, M.: Smoothgrad:
  removing noise by adding noise. arXiv preprint arXiv:1706.03825  (2017)

\bibitem{Springenberg2015StrivingFS}
Springenberg, J.T., Dosovitskiy, A., Brox, T., Riedmiller, M.A.: Striving for
  simplicity: The all convolutional net. CoRR  \textbf{abs/1412.6806} (2015)

\bibitem{srinivas2019fullgrad}
Srinivas, S., Fleuret, F.: Full-gradient representation for neural network
  visualization. In: Advances in Neural Information Processing Systems (2019)

\bibitem{10.5555/3305890.3306024}
Sundararajan, M., Taly, A., Yan, Q.: Axiomatic attribution for deep networks.
  JMLR.org (2017)

\bibitem{thulasidasan2019mixup}
Thulasidasan, S., Chennupati, G., Bilmes, J., Bhattacharya, T., Michalak, S.:
  On mixup training: Improved calibration and predictive uncertainty for deep
  neural networks. arXiv preprint arXiv:1905.11001  (2019)

\bibitem{tsipras2018robustness}
Tsipras, D., Santurkar, S., Engstrom, L., Turner, A., Madry, A.: Robustness may
  be at odds with accuracy (2018), \url{http://arxiv.org/abs/1805.12152}, cite
  arxiv:1805.12152

\bibitem{WahCUB_200_2011}
Wah, C., Branson, S., Welinder, P., Perona, P., Belongie, S.: {The Caltech-UCSD
  Birds-200-2011 Dataset}. Tech. Rep. CNS-TR-2011-001, California Institute of
  Technology (2011)

\bibitem{9150840}
Wang, H., Wang, Z., Du, M., Yang, F., Zhang, Z., Ding, S., Mardziel, P., Hu,
  X.: Score-cam: Score-weighted visual explanations for convolutional neural
  networks. In: 2020 IEEE/CVF Conference on Computer Vision and Pattern
  Recognition Workshops (CVPRW). pp. 111--119 (2020).
  \doi{10.1109/CVPRW50498.2020.00020}

\bibitem{Wang04imagequality}
Wang, Z., Bovik, A.C., Sheikh, H.R., Simoncelli, E.P.: Image quality
  assessment: From error visibility to structural similarity. IEEE Transactions
  on Image Processing  \textbf{13}(4),  600--612 (2004)

\end{thebibliography}

\clearpage
\appendix  

%\documentclass[runningheads]{llncs}
%
%\usepackage[T1]{fontenc}
% T1 fonts will be used to generate the final print and online PDFs,
% so please use T1 fonts in your manuscript whenever possible.
% Other font encondings may result in incorrect characters.
%
%\usepackage{graphicx}
% Used for displaying a sample figure. If possible, figure files should
% If you use the hyperref package, please uncomment the following two lines
% to display URLs in blue roman font according to Springer's eBook style:
%\usepackage{color}
%\renewcommand\UrlFont{\color{blue}\rmfamily}
%
%\usepackage{todonotes}

%\begin{document}
%
\section*{Supplementary Material}

\section{Interpretation method details}
\subsubsection{Model aware interpretation methods}
The \emph{Sensitivity} saliency map  shows the sensitivity of the output to infinitesimal changes in each input feature. It is obtained by simply taking the derivative of the score w.r.t to the input :
\begin{equation}
S_{x, F}\left(x_{i}\right) =\sum_{ channels}\frac{\partial F(x)_{c}}{\partial {x}} (x_{i}) 
\end{equation}

% \subsubsection{Integrated Gradient}
\emph{Integrated Gradient} improves on the Sensitivity method and avoids the null gradient issue by computing the average of gradients obtained for images along the convex path created by the interpolation between the input image and a reference image. Given $x^{r}$ the reference image, the formula used practically is :

\begin{equation}
S_{x, F}\left(x_{i}\right)=  \frac{ \left(x_{i}-x^{r}_{i}\right) }{m}  \sum_{k=1}^{m}\frac{\partial
F\left(x^{r}_{i}+ \frac{k}{m}\left(x_{i}-x^{r}_{i}\right)\right)}{\partial {x}}\,(x_{i})
\end{equation}

% \subsubsection{Non-gradient model based interpretability}
Finally, \emph{LayerWise Relevance Propagation} (LRP) was used. It belongs to the family of interpretation algorithms that leverage the inside architecture of the model but without relying on the gradients. It backpropagates the output score $F(x)_{c}$ through the model through a signal called relevance.
The main rule is the conservation of the relevance between layers, the sum of all relevances of a layer should be equal to the output score : 
\begin{equation}
\forall\ { l},  F(x)_{c} \approx \sum_{p} R_{p}^{(l)}
\end{equation}
To do so the score is backpropagated from one neuron to another proportionally to each neuron activation and to the strength of the link between them.
\begin{equation}
R^{l}_{j}=\sum_{k} \frac{a_{j} w_{j k}}{\epsilon+\sum_{0, j} a_{j} w_{j k}} R^{l+1}_{k}
\end{equation}
with $w_{j,k}$ the weight between the neuron j of the l-th layer and k, $a_{j}$  the activation of the $j$-th neuron of the l+1-th layer and $\epsilon$ a jitter parameter which stabilizes numerically the propagation. 

\subsubsection{Model agnostic interpretation methods}

We also used methods that consider models as black-boxes such as occlusion-based approaches which degrade the input and analyse predicted score variations to define the importance of each part of the input. We used RISE and Meaningful Perturbation in our experiments.

\emph{RISE}'s saliency map  is obtained by computing the sum  of randomly sampled masks weighted by the scores obtained when the model is applied to these masked image. Given a certain mask $m_{j}$ and the number of samples $N$, we have :
\begin{equation}
S_{x, F}\left(x_{i}\right) = \frac{1}{\mathbb{E}\left([m]\right)*N}\sum_{j=1}^{N}F(I\odot m_{j})(x_{i})
\end{equation}

\emph{Meaningful Perturbation} (MP) tries to find the smallest mask which makes the prediction score drop the most by optimizing :

\begin{equation}
\min _{m \in[0,1]^{H*W}} \lambda\|1-m\|_{1} + \beta TV(m) +F(x\odot m)_{c}
\end{equation}

with $TV$ being the total variation of the mask. 
This parameter penalizes the shape of the mask to be as regular as possible in order to avoid adversarial artifacts.

\end{document}